\documentclass[10pt,twocolumn,letterpaper]{article}

\usepackage{iccv}
\usepackage{times}
\usepackage{epsfig}
\usepackage{graphicx}
\usepackage{amsmath}
\usepackage{amssymb}
\usepackage{mathtools}

\usepackage[nottoc]{tocbibind}
\usepackage[utf8]{inputenc} 
\usepackage[T1]{fontenc}    
\usepackage{url}            
\usepackage{multirow}
\usepackage{booktabs}       
\usepackage{amsfonts}       
\usepackage{nicefrac}       
\usepackage{microtype}      
\usepackage{subcaption}
\usepackage{xcolor}
\usepackage{array}
\usepackage[T1]{fontenc}
\usepackage{fix-cm}
\usepackage{bm}
\usepackage{float}

\graphicspath{ {figcvpr/} }

\newcolumntype{K}[1]{>{\centering\let\newline\\\arraybackslash\hspace{0pt}}m{#1}}

\newcommand{\specialcell}[2][c]{%
  \begin{tabular}[#1]{@{}c@{}}#2\end{tabular}}
  
\usepackage{tikz}
\newcommand{\mypm}{\mathbin{\tikz [x=1.4ex,y=1.4ex,line width=.1ex] \draw (0.0,0) -- (1.0,0) (0.5,0.08) -- (0.5,0.92) (0.0,0.5) -- (1.0,0.5);}}%

\usepackage[pagebackref=true,breaklinks=true,letterpaper=true,colorlinks,bookmarks=false]{hyperref}

\iccvfinalcopy 

\def\iccvPaperID{395} 

\ificcvfinal\fi
\begin{document}

\title{Improving Robustness of Feature Representations to Image Deformations using Powered Convolution in CNNs}

\author{Zhun Sun, Mete Ozay, Takayuki Okatani\\
Tohoku University, Japan \\
Institution1 address\\
{\tt\small sun,mozay,okatani@vision.is.tohoku.ac.jp}
}

\maketitle

\begin{abstract}
In this work, we address the problem of improvement of robustness of feature representations learned using convolutional neural networks (CNNs) to image deformation. We argue that higher moment statistics of feature distributions could be shifted due to image deformations, and the shift leads to degrade of performance and cannot be reduced by ordinary normalization methods as observed in experimental analyses. In order to attenuate this effect, we apply additional non-linearity in CNNs by combining power functions with learnable parameters into convolution operation. In the experiments, we observe that CNNs which employ the proposed method obtain remarkable boost in both the generalization performance and the robustness under various types of deformations using large scale benchmark datasets. For instance, a model equipped with the proposed method obtains 3.3\% performance boost in mAP on Pascal Voc object detection task using deformed images, compared to the reference model, while both models provide the same performance using original images. To the best of our knowledge, this is the first work that studies robustness of deep features learned using CNNs to a wide range of deformations for object recognition and detection.
\end{abstract}
\vspace{-0.2cm}

\section{Introduction}
\label{sect_intro}
Recognition of objects using deformed images is a challenge that has been studied extensively in computer vision and pattern recognition in the last decade~\cite{bay2006interactive, chechik2008max, Garcia-Laencina2010, leonardis2000robust, lindner2004robust, marlin2008missing, 10.3389/fpsyg.2013.00124}. While convolutional neural networks (CNNs) have achieved impressive progress for object classification and recognition in some benchmark datasets~\cite{he2015deep, ioffe2015batch, szegedy2015going, simonyan2014very}, recent works~\cite{da2016empirical, dodge2016understanding} show that their performance is severely degraded for deformed images. 

\begin{figure}[t]
\centering
    \begin{minipage}[c]{0.21\textwidth}
        \captionsetup[subfigure]{justification=centering}
        \centering
        \subcaption*{\textbf{Original Images}}
    \end{minipage}
    \begin{minipage}[c]{0.21\textwidth}
        \captionsetup[subfigure]{justification=centering}
        \centering
        \subcaption*{\textbf{Deformed Images}}
    \end{minipage} \\
    \vspace{-0.1cm}
    
    \begin{minipage}[c]{0.21\textwidth}
        \includegraphics[width=1\textwidth]{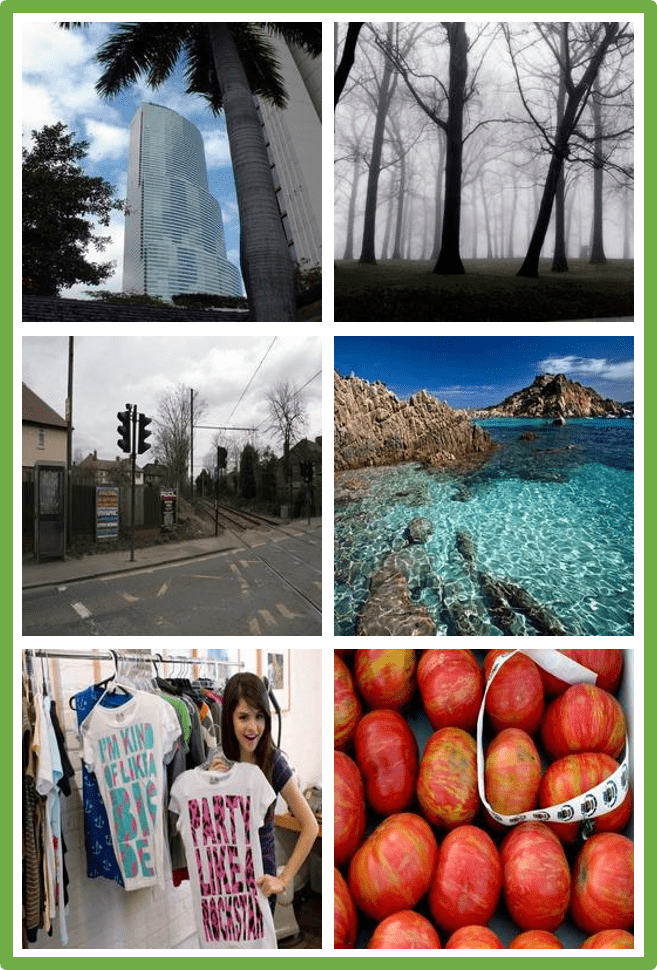}
    \end{minipage}
    \begin{minipage}[c]{0.21\textwidth}
        \includegraphics[width=1\textwidth]{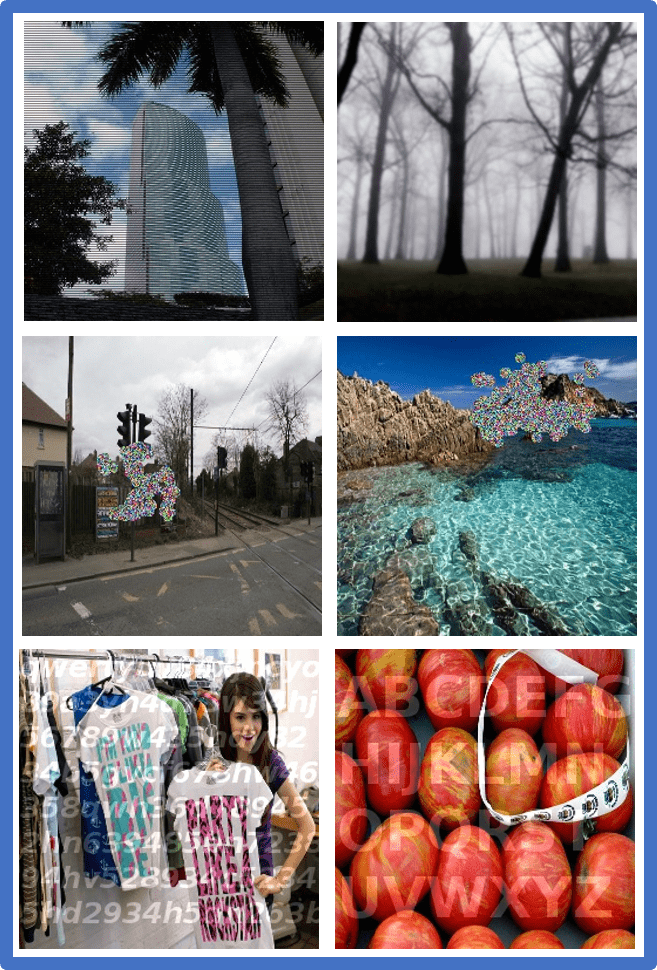}
    \end{minipage} \\
   
    \begin{minipage}[c]{0.40\textwidth}
    \centering
        \includegraphics[width=1\textwidth,height=0.42\textwidth]{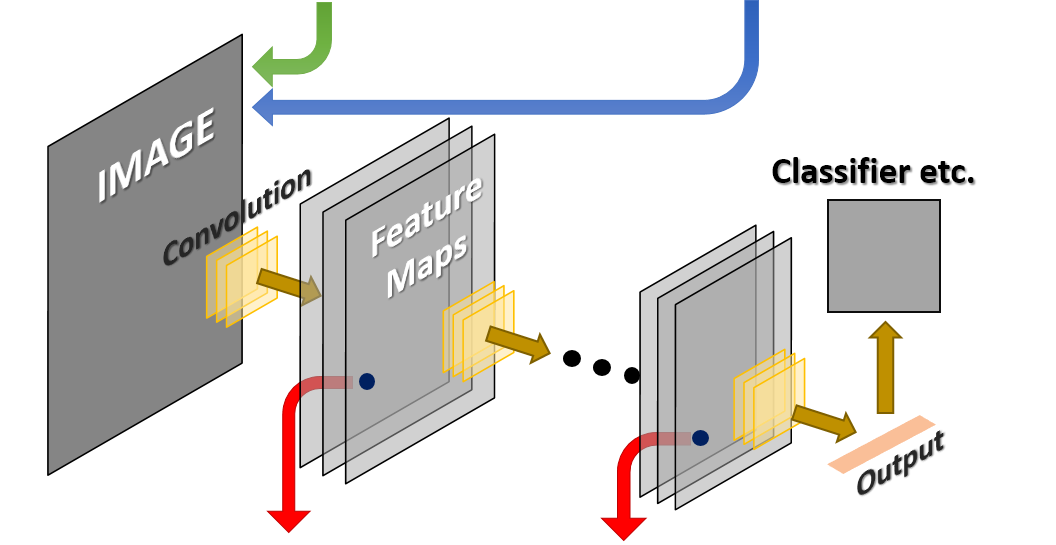}
    \end{minipage} \\
    
    \begin{minipage}[c]{0.20\textwidth}
        \includegraphics[width=1\textwidth,height=0.8\textwidth]{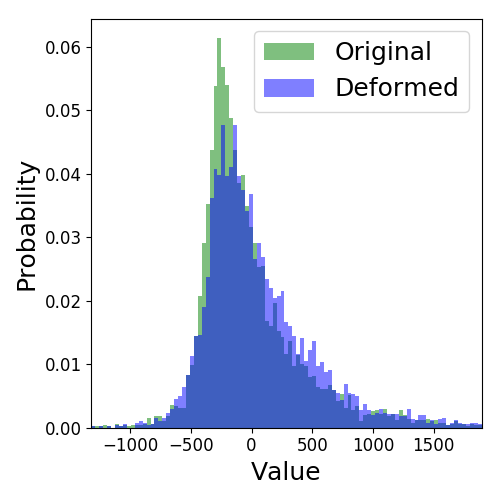}
    \end{minipage}
    \begin{minipage}[c]{0.20\textwidth}
        \includegraphics[width=1\textwidth,height=0.8\textwidth]{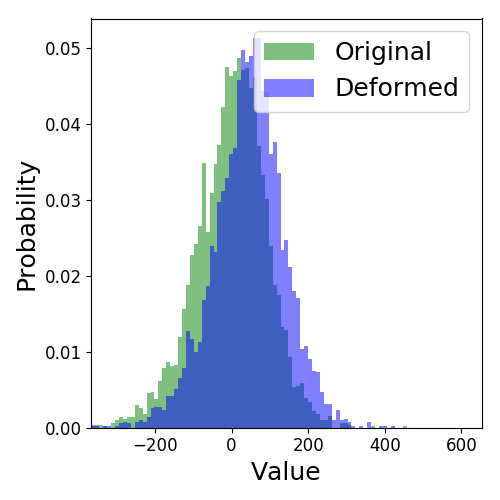}
    \end{minipage}
    \begin{minipage}[c]{0.02\textwidth}
        \subcaption*{\textbf{~}}
    \end{minipage} \\
    
\caption{Divergence between distributions of neuron activities can be observed given original and deformed input images. This effect is accumulated through layers of CNNs, and could finally result in degraded performance. The probability densities are calculated based on 5,000 original and manually deformed images from validation set of ILSVRC-12 using VGG-16~\cite{simonyan2014very}. More results are provided in supplementary material.}
\vspace{-0.6cm}
\label{fig1}
\end{figure}

In this work, we consider a collection of deformations that are observed in real world datasets of natural images in object recognition and detection tasks. Specifically, we consider the following deformation types: i) geometric transformations, ii) signal processing loss, iii) statistical noise and iv) occlusion (see Section~\ref{subsect_robust} for details). Image deformations result in change of statistical properties of datasets. In other words, recognition of objects using deformed datasets can be posed as a dataset shift problem. Suppose that $\mathbf{X_{tr}}$ and $\mathbf{X_{te}}$ are sets of features detected using datasets $\mathbf{D_{tr}}$ and $\mathbf{D_{te}}$, respectively. Moreover, suppose that a CNN model is trained using $\mathbf{D_{tr}}$, and $\mathbf{D_{te}}$ is a set of deformed data. Then, the deformation (such as noisy pixels) occurred on the dataset $\mathbf{D_{te}}$ will influence detection of feature representations learned using $\mathbf{D_{tr}}$. Consequently, we may observe that feature distributions $p(\mathbf{X_{tr}})$ and $p(\mathbf{X_{te}})$ diverge.

Feature normalization is a popularly employed solution of the aforementioned problem, which is helpful in minimizing divergence concerning mean and variance. However, normalization methods do not ensure minimization of divergence of higher moments, e.g. skewness and kurtosis. The significance of higher moments in performance of several vision tasks has been studied in the last decade, such as for analysis of change of judgment of surface reflectance by manipulation of these statistical properties~\cite{Sharan-JOSA-08}. In CNNs, massive shifts of higher moments are usually observed when test images are deformed (Figure~\ref{fig1}), together with a decrement of performance (see Section~\ref{quick_glance}). 

However, higher moments of a distribution can not be easily normalized like mean and variance, an alternative approach is required in order to resolve the aforementioned problem. Inspired by the fact that, CNNs are complicated functions that map the distribution of given inputs into desired spaces of distribution (e.g. the probabilities of classes in a classification task), by stacking linear and non-linear transformation repeatedly. We consider the approach that to estimate mappings such that the divergence between distributions employed in the new spaces can be reduced, by applying additional non-linearity to CNNs. To this end, in this paper we employ power functions with learnable exponents to estimate non-linear mappings between distributions, where the exponents can be estimated by minimizing loss functions of CNNs. Our contributions can be summarised as follows:
\begin{enumerate}
    \item We explore that the shift of higher moments of feature distributions could lead to a performance degrade. In addition, we investigate the viability of reducing divergence by using normalization methods and non-linear functions.
    \item We propose a new method to apply non-linearity by integrating the power function and the convolution operation. In addition, we propose a practical method to train parameters of power functions, such that the distribution divergence problem can be attenuated without causing a vanishing gradient problem. 
    \item In experimental analyses, we demonstrate that the generalization performance of CNNs, and their robustness to various types of image deformations can be substantially improved using our proposed method, for object recognition and detection tasks using large scale datasets (ILSVRC-12, Places2 and Pascal Voc).
\end{enumerate}


\subsection{Related Work}
\label{related work}
\noindent \textbf{Processing Deformed Images using/in Deep NNs}: The dataset shift problem caused by image deformation has been previously tackled using several approaches~\cite{kanazawa2014locally, sohn2012learning, tang2010deep}. The most widespread approach used to minimize the divergence, is to employ a generative model $p(\mathbf{z})p_{\mathbf{\theta}}(\mathbf{X}|\mathbf{z})$ such that both $p(\mathbf{X_{tr}})$ and $p(\mathbf{X_{te}})$ could be inferred from a fixed distribution $p(\mathbf{z})$ which is parametrized by a set of parameters $\mathbf{\theta}$~\cite{hinton2011transforming, 2011arXiv1104.4153R, zou2012deep}. Due to the intractability of $p_{\mathbf{\theta}}(\mathbf{X}|\mathbf{z})$, it could be difficult to estimate parameters $\mathbf{\theta}$~\cite{lenc2015understanding}. Recent works~\cite{Cheng_2016_CVPR, jaderberg2015spatial} have considered modeling of some specific transformation patterns such as scale and rotation by learning sub-networks with a parameter set $\phi$. These sub-networks could yield a new distribution for testing data $q_{\phi}(\mathbf{X_{te}})$ that is similar to $p(\mathbf{X_{tr}})$, or a new $q_{\phi}(\mathbf{X_{tr}})$ that is likely to be an approximation to $p(\mathbf{X_{te}})$. Still, estimation of parameters $\mathbf{\phi}$ is difficult since the transformation patterns could be very different for $\mathbf{X_{tr}}$ and $\mathbf{X_{te}}$. 

\noindent \textbf{Normalization Methods}: State-of-the-art normalization methods such as Batch Normalization (BN)~\cite{ioffe2015batch} and Layer Normalization (LN)~\cite{2016arXiv160706450L} are considered to be able to reduce the inherent and data-wise shift problem by fixing the mean and the variance of distribution of input features within each layer. Concretely, when receiving $\mathbf{X}$ (which denotes either a data matrix of train/testing samples, or a matrix of features extracted from samples), normalization methods output $\tilde{X} = \frac{\mathbf{X} - \mathbf{\mu_X}}{\mathbf{\sigma_X}}$ that has zero mean and unit variance, by subtracting mean $\mathbf{\mu_X}$ and standard deviation $\mathbf{\sigma_X}$. Although these normalization methods are confirmed to work well empirically, still they assume that data are distributed with the same parametrized function such as Gaussian, without considering the classification loss. However this assumption does not apply to general cases as mentioned above, especially when the divergence usually obtained from higher moment statistics, i.e. skewness and kurtosis. In practice, these normalization methods are followed by a linear transformation. This ad-hoc method slightly mitigates the problem, but still there remains large divergence between $p(\mathbf{X_{tr}})$ and $p(\mathbf{X_{te}})$.


\noindent \textbf{Use of Power Functions}: Although various types of non-linearity functions have been explored and proved to be beneficial for training CNNs empirically, functions endowed with a power operation have been barely utilized. A former attempt~\cite{GulcehreCPB13} proposed an $\ell_p$ nonlinear unit with a learnable order $p$, which receives signals from several projections of a subset of units of the previous layer, and performs $\ell_p$ normalization. This can be interpreted as an implicit employment of power operation which could estimate and assign different weights to different feature activations, and their results show performance improvement in some benchmark datasets. S-shaped rectified linear units (SReLU) \cite{JinXFWXY15} have been proposed to achieve more complicated non-linearity that are considered to be able to imitate the behavior of power or logarithm functions with performance boost.



\section{Our Proposed Approach}
\label{pow_func}
Consider a convolution kernel with $D$ output channels $\mathbf{W} \in \mathbb{R}^{C \times D \times h \times w}$ that slides on a tensor of features $\mathbf{T} \in \mathbb{R}^{C \times H \times W}$, then the output of convolution operation $\mathbf{U} \in \mathbb{R}^{D \times \tilde{H} \times \tilde{W}}$ can be computed by 
\vspace{-0.1cm}
\begin{equation}
    \mathbf{U}_d = \mathbf{W} \otimes \mathbf{T} = \sum_{c \leq C} {\mathbf{W}_{c,d} \star \mathbf{T}_c },
\end{equation}
\vspace{-0.1cm}
where $\star$ is the two dimensional convolution operation. In order to append power function into the convolution operation, we suggest two methods, called, in-channel power convolution and out-channel power convolution, which are defined by
\vspace{-0.1cm}
\begin{equation}
    \mathbf{U}_d^{I} = \sum_{c \leq C} {\mathbf{W}_{c,d} \star \psi(\mathbf{T}_c, \alpha_{c, d}, \beta_{c, d}) }, 
\end{equation}
\vspace{-0.1cm}
\begin{equation}    
    \mathbf{U}_d^{O} = \sum_{c \leq C} {\psi(\mathbf{W}_{c,d} \star \mathbf{T}_c, \alpha_{c, d}, \beta_{c, d}) },
\end{equation}
\vspace{-0.1cm}
where $c = 1, 2, \ldots , C$ and $d = 1, 2, \ldots , D$ denotes the index of input and output channels respectively. $\psi$ is the element-wise power function operation defined by the following equation, $\alpha_{c, d}, \beta_{c, d} \in \mathbb{R}$ are the corresponding channel-wise. 
\begin{equation}
\label{eq4}
    \mathbf{y} = \psi(\mathbf{x}, \alpha, \beta) = 
    \begin{dcases}
        \frac{x_m^{\alpha+1}}{\beta+1},   & \text{if } x_m \geq 0\\
        -\frac{(-x_m)^{\alpha+1}}{\beta+1},   & \text{otherwise}
    \end{dcases},
\end{equation}
where $x_m$ is the $m^{th}$ element of $\mathbf{x}  \in \mathbb{R}^{M}$, $\alpha$ and $\beta$ are trainable exponent and scale parameters. Since power function is defined on $\mathbb{R}^+$, we apply a mirror for the negative inputs (the exponent and scale parameters can also differ from those employed for positive inputs). If the network employs activation function that only output positive values, such as ReLU, then the negative inputs can be safely ignored for in-channel power convolution. 

In practice, offering different sets of $\alpha$ and $\beta$ for all the channels is not sensible, since $c$ and $d$ usually take large values in DNNs, which results in an enormous increment of parameters and computational complexity. Therefore, we suggest two methods to share $\alpha$ and $\beta$ among channels. Concretely, for in-channel power convolution, we split $D$ output channels into $\Lambda$ portions and all the channels within $D_\lambda$ share the same set of $\alpha_{c, D_\lambda}, \beta_{c, D_\lambda}$ ($\lambda = 1, ... ,\Lambda$). For out-channel power convolution, we also employ shared sets of parameters $\alpha_{C_\lambda, d}, \beta_{C_\lambda, d}$. Especially if $\Lambda = 1$, that is, all the in- and out- channels share the same set of parameters, thus $\mathbf{U}^{I}$ and $\mathbf{U}^{O}$ would make no difference. 

The parameters $\alpha$ and $\beta$ are estimated using gradients computed during back-propagation (BP) in a DNN by ($d$ and $c$ are omitted for simplicity)
\vspace{-0.1cm}
\begin{equation}
\label{equation3}
    \frac{\partial \mathcal{L}} {\partial \alpha} = \sum_{i=1}^m \frac{\partial \mathcal{L}} {\partial y_i} y_i  \ln |x_i| ,
\end{equation} 
\begin{equation}
    \frac{\partial \mathcal{L}} {\partial \beta} = \sum_{i=1}^m - \frac{\partial \mathcal{L}} {\partial y_i} \frac{y_i}{\beta+1} ,
\end{equation} 
\begin{equation}
\label{equation4}
    \frac{\partial \mathcal{L}} {\partial x_i} =
    sgn(x_i) \frac{\partial \mathcal{L}} {\partial y_i} \frac{\alpha+1}{\beta+1}|x_i|^{\alpha+1} =  (\alpha+1) \frac{\partial \mathcal{L}} {\partial y_i} \frac{y_i}{x_i} ,
\end{equation} 
where $sgn$ is the sign function and $\mathcal{L}$ denotes a loss function of the DNN such as classification loss. Note that $x_i$ should not be $0$, otherwise we assign a $0$ to the gradient. 

According to \eqref{equation4}, the error obtained from $\frac{\partial \mathcal{L}}{\partial y_i}$ during back-propagation will be amplified/reduced depending on the quotient of input and output data $\frac{y_i}{x_i}$. Usually this is not a problem unless $\alpha \ll 0$ or $\alpha \gg 0$, where we may observe a possible exploding/vanishing gradient problem. Especially, $\alpha \approx -1$ will give an output close to $1$ for any input. Thus, an input with \textit{small} magnitude of activation will provide \textit{large} error during back-propagation, and vice versa. Consequently, training progress could be unstable such that they either diverge or stop to further converge. In order to address this problem, algorithmically, we employ $\ell_2$ regularized terms towards $\alpha$ and $\beta$ for computation of the final loss of power convolution during training. For the models that are trained using large learning rate and small regularization coefficients, we further employ a variable learning rate scheme that shrinks the gradients by multiplying $\cos(\pi \alpha /2)$ when $\alpha \ll 0$ or $\alpha \gg 0$.



\begin{figure*}[t]
\centering
    \begin{minipage}[c]{0.195\textwidth}
        \includegraphics[width=1\textwidth]{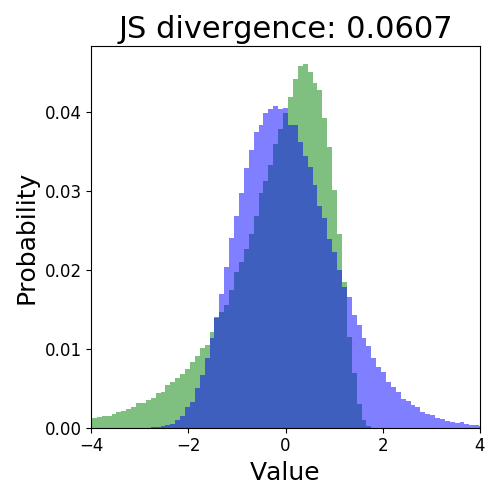}
    \end{minipage}
    \begin{minipage}[c]{0.195\textwidth}
        \includegraphics[width=1\textwidth]{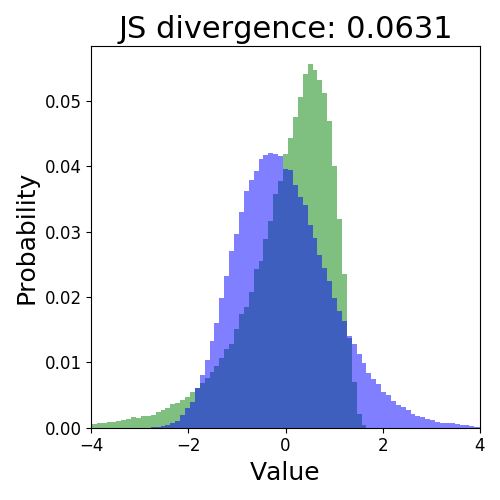}
    \end{minipage}
    \begin{minipage}[c]{0.195\textwidth}
        \includegraphics[width=1\textwidth]{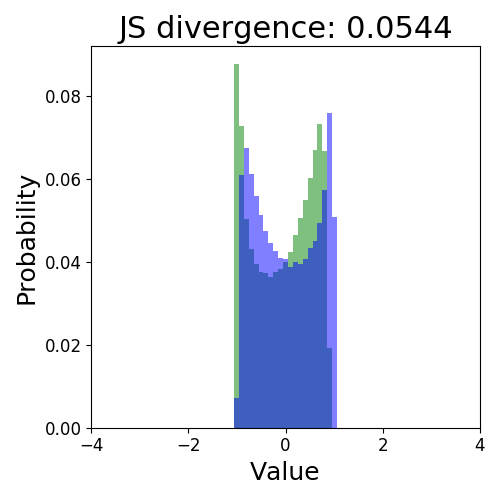}
    \end{minipage} 
    \begin{minipage}[c]{0.195\textwidth}
        \includegraphics[width=1\textwidth]{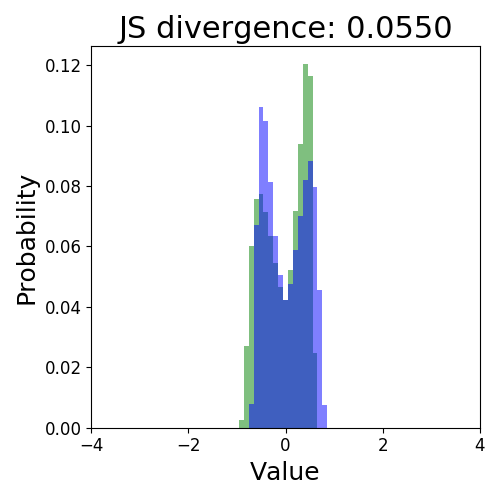}
    \end{minipage} 
    \begin{minipage}[c]{0.195\textwidth}
        \includegraphics[width=1\textwidth]{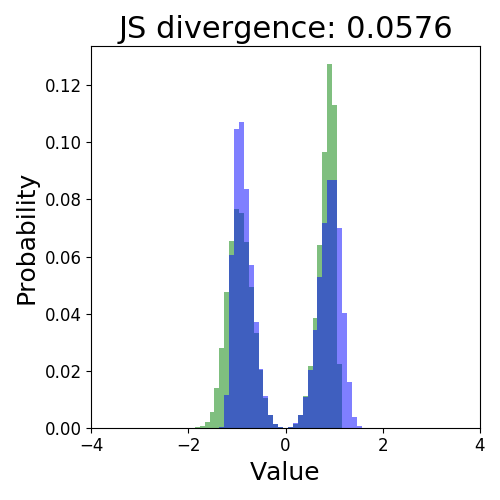}
    \end{minipage}
    \\
    
    \begin{minipage}[c]{0.195\textwidth}
        \captionsetup[subfigure]{justification=centering}
        \centering
        \subcaption{Original}
    \end{minipage}
    \begin{minipage}[c]{0.195\textwidth}
        \captionsetup[subfigure]{justification=centering}
        \centering
        \subcaption{Normalization}
    \end{minipage}
    \begin{minipage}[c]{0.195\textwidth}
        \captionsetup[subfigure]{justification=centering}
        \centering
        \subcaption{Tanh}
    \end{minipage} 
    \begin{minipage}[c]{0.195\textwidth}
        \captionsetup[subfigure]{justification=centering}
        \centering
        \subcaption{Softsign}
    \end{minipage}
    \begin{minipage}[c]{0.195\textwidth}
        \captionsetup[subfigure]{justification=centering}
        \centering
        \subcaption{Power (Exp.$=0.25$)}
    \end{minipage}
    \\
    \vspace{-0.4cm}

\caption{A demonstration of changes of distributions that are observed by using different nonlinear functions. In the analyses, $\kappa$ value of (a) original blue and green distribution was set to $-0.2$ and $0.5$, respectively. (b)(c)(d)(e) depict distributions computed after employment of the corresponding transformations. It can be seen that distributions mapped by use of additional non-linearity have smaller divergence with better fitted shapes for this task.}
\vspace{-0.4cm}
\label{fig2}
\end{figure*}

\subsection{An Analysis of Non-linearity}
\label{quick_glance}

As discussed in Section~\ref{sect_intro}, divergence caused by shifted skewness and kurtosis between feature distributions, is harmful in inference using new samples. However, minimization of this type of divergence cannot be achieved by normalization methods (see Section~\ref{related work}). Thus we consider an alternative approach, by introducing additional non-linearity to CNNs, space of diverged distributions can be mapped to a new space in which the divergence between distributions could be minimized. More precisely, we say that the feature distributions $p(\mathbf{X_{tr}})$ and $p(\mathbf{X_{te}})$ are mapped into new distributions $\tilde{p}_{\theta}(\mathbf{X_{tr}})$ and $\tilde{p}_{\theta}(\mathbf{X_{te}})$, such that we obtain $d(\tilde{p}_{\theta}(\mathbf{X_{te}}) || \tilde{p}_{\theta}(\mathbf{X_{tr}})) \leq d(p(\mathbf{X_{te}}) || p(\mathbf{X_{tr}}))$, where $d$ is a function that measures similarity between distributions such as Jensen–Shannon divergence. 

In order to illustrate this, we design a multi-class classification experiment using a two-layer neural network that has 128 hidden units. We employed ReLU after the hidden layer for all models. The inputs are $M$-dimensional vectors $\mathbf{X} \in \mathbb{R}^M$. Each vector $\mathbf{X}$ consists of $N < M$ features that are used to identify class labels, and the rest $M-N$ features are utilized as noise. All variables of $\mathbf{X}$ are sampled from a generalized normal distribution defined by~\eqref{gnd} at position $\xi = 0$ and scale $\alpha = 1$. 
\begin{equation}
\label{gnd}
    \frac{\phi(y)}{\alpha-\kappa(x-\xi)}, y = \begin{dcases} - \frac{1}{\kappa} \log \left[ 1- \frac{\kappa(x-\xi)}{\alpha} \right] & \text{if } \kappa \neq 0 \\ \frac{x-\xi}{\alpha} & \text{if } \kappa=0 \end{dcases}
\end{equation}

\setlength{\tabcolsep}{5.2pt}
\begin{table}[t]
  \centering
  \begin{tabular}{ccccc}
    \toprule
    Total Features & 
    \multicolumn{4}{c}{\textbf{128}} \\
    \cmidrule(lr){2-5}
    Determinant Features &
    1 & 2 & 4 & 8 \\
    \midrule
    Base w/o divergence & 
    96.4 & 92.6 & 83.0 & 36.3 \\
    Base & 
    93.9 & 87.9 & 73.7 & 26.9 \\
    \midrule
    Batch Norm. Only &
    92.1 & 84.9 & 71.5 & 29.3 \\
    Batch Norm. + Trans. &
    95.2 & 89.2 & 77.5 & 39.1 \\
    Base + 1 layer &
    93.2 & 85.1 & 70.5 & 27.8 \\
    Base + 2 layers &
    93.1 & 85.3 & 69.6 & 29.8 \\
    Tanh &
    97.4 & 95.1 & 89.4 & 59.5 \\
    Softsign &
    98.4 & 96.9 & 93.0 & 65.7 \\
    \textbf{Power} & 
    \textbf{99.0} & \textbf{98.4} & \textbf{97.5} & \textbf{77.6} \\
    \bottomrule
  \end{tabular}
  \caption{Averaged classification accuracy (\%) of two-layer neural network models using artificial datasets over 10 runs. Trans. stands for the followed linear transformation proposed as BN~\cite{ioffe2015batch}. More results will be provided in supplementary materials.}
  \vspace{-0.5cm}
  \label{table1}
\end{table}

We randomly choose the shape parameter $\kappa$ from a uniform distribution $\mathcal{U}(-1,1)$ from all the features. The corresponding labels are evaluated by ${\sum_{n=0}^{N-1}2^n \cdot \mathbf{1}(X_n\geq 0)}$, where $\mathbf{1}(\cdot)$ is an indicator function that outputs $1$ when the argument of the indicator is true. Then we have $2^N$ number of classes, as $N$ grows, it is more difficult for networks to distinguish the useful features during training. We choose $M = 128$ and ${N \in \{1, 2, 4, 8\}}$ in the experiment, 10,000 training and 10,000 test samples are generated individually for each configuration. 

Initially, we first generate a set of data using the same shape parameter $\kappa$ for both training and test sets as a reference set (Base w/o divergence). Then, we generate diverged datasets using different $\kappa$ to construct both training and test data (Base). We first compare the performance of the base two-layer neural network trained on both of them. Then, we employ different non-linearity (on input vectors), and test the performance using diverged datasets. The results are given in Table~\ref{table1}. It can be seen that, if the distributions of features is shifted by higher moment statistics, then the performance of base model (Base) is degraded notably in all cases. While BN seems to be helpful, normalization without using linear transformation even performs worse than Base, except the cases where the number of classes is large. Furthermore, in Figure~\ref{fig2}, we can see that the divergence is even larger for the normalized data compared to original data. The results indicate that linear transformation contributes to improvement of the robustness to divergence more than normalization. 

In the experiments that implement non-linearity on input vectors, we evaluate the change of performance using power function with trainable exponents and scale parameters defined in~\eqref{eq4} together with two reference functions, Tanh~\cite{lecun2012efficient} and  Softsign~\cite{Bergstra+2009}. We can also consider that implementation of deeper NNs with larger number of layers as an implicit use/computation of additional non-linearity. It can be seen that, with additional non-linearity, the neural networks overcome the shift in distributions by a large margin, although a decrement of accuracy appears in the models endowed with more layers (discussed in Section~\ref{expl_power}). Again, note that, employment of additional non-linearity is not targeting at removing skewness and kurtosis, but rather mapping them into distributions that conditionally better and less diverged (Figure~\ref{fig2}).

Here we choose to implement power functions, not only because it outperformed all the other methods with a substantial margin, but also Tanh and Softsign are no longer commonly utilized in state-of-the-art DNNs (have been replaced by rectified non-linearity~\cite{nair2010rectified}) due to the vanishing gradient problem, which can be resolve for power using the training tricks mentioned in last section. 


\begin{table}[t]
  \centering
  \begin{tabular}{cccc}
    \toprule
    \multirow{2}[4]{*}{\specialcell{\textbf{Network} \\ \textbf{Size}}} & \multicolumn{3}{c}{\textbf{Dimension of Input}}\\
    \cmidrule(lr){2-4}
     & \textbf{16} & \textbf{64} & \textbf{256}\\
    \midrule
    \textbf{64} & $24.6\mypm 18.4$ & $38.2\mypm 27.4$ & $40.2\mypm 28.3$ \\
    \textbf{256} & $20.0\mypm 14.4$ & $36.0\mypm 26.0$ & $40.9\mypm 29.4$ \\
    \textbf{1024} & $17.7\mypm 13.0$ & $32.2\mypm 23.1$ & $41.1\mypm 29.0$ \\
    \bottomrule
  \end{tabular}
  \caption{Experiments on artificial datasets. The set of ratio $\mathcal{R}$ (mean $\pm$ standard deviation) of average error and the real output is given for each input in percent~(\%). Smaller $\mathcal{R}$ obtain for a network indicates that the network gives a better approximation of power operation. }
  \vspace{-10pt}
  \label{table2}
\end{table}

\subsection{Efficiency of Power Functions as Additional Non-linearity} 
\label{expl_power}
Another practical motivation for explicit employment of power operation is computational burden of its implementation in CNNs. First, it is not straightforward to approximate power operations for a sub-network without using \textit{sufficient} training samples. Second, size of such a sub-network cannot be increased easily, due to computational cost and apprehension of over-fitting (Table~\ref{table1}). To demonstrate this, we design another artificial experiment that aims to mimic power operations using neural networks. We generate a pair of $M$-dimensional vectors $\mathbf{x}, \bm{\alpha} \in \mathbb{R}^M$ whose elements are sampled from $\mathcal{U}(0,1)$ and $\mathcal{U}(0,2)$ respectively ($\mathcal{U}(a,b)$ is the uniform distribution on the interval $[a,b]$). We feed $\mathbf{x}$ and $\mathbf{y}$ ($y_m = x_m^{\alpha_m}, m=1,\ldots,M$) into a two-layer neural network with $N$ hidden units and optimize the network by minimization of the mean squared error.

We use the ratio between averaged error and the real output for each input to evaluate the performance of a network on approximating power functions, by calculating $R_i =\frac{{\sum_{m=1}^M |Y_{m,i,pred}-Y_{m,i,test}|}}{{\sum_{m=1}^M Y_{m,i,test}}}$, where $i$ is the index of samples. We choose $M \in \{16, 64, 256\}$ and $N \in \{64, 256, 1024\}$. For each configuration, we generate $I = 10000$ training and testing samples, respectively, and repeat the experiment 10 times. The \textit{best} results obtained for each configuration are given in Table~\ref{table2}. Obviously, it is cumbersome to estimate negligible number of $\bm{\alpha}$, by employing a network with even hundred thousands of parameters. Especially when the dimension of input grows, increasing the number of units used in a hidden layer seems not to be helpful in obtaining better approximation. Given the fact that, in the state-of-the-art CNN architectures~\cite{he2015deep, simonyan2014very, szegedy2015going}, feature maps may contain several hundreds of output channels. In other words, the network needs to approximate several hundreds of $\bm{\alpha}$ for each feature map, in order to reproduce the non-linearity provided by power functions, which is unachievable as suggested in the experiment. 

\section{Experimental Results}
\label{exp_results}

\begin{table}[t]
  \centering
  \begin{tabular}{ccc}
    \toprule
    \textbf{Models} & \textbf{Cifar-10} & \textbf{Cifar-100}    \\
    \midrule
    \small
    Base & $21.80 \mypm 0.29$ & $51.50 \mypm 0.36$ \\
    IN-ch. ($\Lambda=1$) & $\textbf{20.48} \mypm \textbf{0.37}$ & $\textbf{50.21} \mypm \textbf{0.44}$ \\
    IN-ch. ($\Lambda=2$) & $21.33 \mypm 0.50$ & $51.08 \mypm 0.54$ \\
    IN-ch. ($\Lambda=4$) & $21.29 \mypm 0.25$ & $51.25 \mypm 0.40$ \\
    IN-ch. ($\Lambda=8$) & $21.31 \mypm 0.38$ & $50.98 \mypm 0.38$ \\
    OUT-ch. ($\Lambda=2$) & $21.24 \mypm 0.29$ & $50.88 \mypm 0.57$ \\
    OUT-ch. ($\Lambda=4$) & $22.48 \mypm 0.39$ & $52.36 \mypm 0.29$ \\
    OUT-ch. ($\Lambda=8$) & $24.39 \mypm 0.37$ & $53.38 \mypm 0.25$ \\
    \midrule
    ResNet-36 & 7.76 & 31.57 \\
    ResNet-36-IN-ch. ($\Lambda=1$) & 7.55 & 31.12 \\
    ResNet-36-IN-ch. ($\Lambda=2$) & 7.66 & 30.78 \\
    ResNet-36-IN-ch. ($\Lambda=4$) & \textbf{7.45} & \textbf{30.45} \\
    \bottomrule
  \end{tabular}
  \caption{Classification error on Cifar-10/100 for different configurations.}
  \vspace{-10pt}
  \label{table3}
\end{table}

\begin{table}[t]
  \centering
  \begin{tabular}{cccc}
    \toprule
    \multirow{2}[0]{*}{\textbf{Models}} & 
    \multirow{2}[0]{*}{\specialcell{\textbf{Error(\%)}\\ \textbf{Top1/Top5}}} &
    \multirow{2}[0]{*}{\specialcell{\textbf{Training}\\ \textbf{Time(ms)}}} &
    \multirow{2}[0]{*}{\specialcell{\textbf{Number}\\ \textbf{Parameters}}} \\ 
    \\
    \midrule
    Base & 37.3/14.9 & 295 ms & $\approx$7.4M \\
    IN-ch. ($\Lambda=1$) & 35.9/14.1 & 327ms & $\approx$1.5K more \\
    \textbf{IN-ch. ($\Lambda=2$)} & \textbf{35.4/13.8} & 399 ms & $\approx$3K more \\
    \midrule
    Base & 48.6/17.6 & 294ms & $\approx$7.4M \\
    IN-ch. ($\Lambda=1$) & 47.8/17.3 & 326ms  & $\approx$1.5K more \\
    \textbf{IN-ch. ($\Lambda=2$)} & \textbf{47.4/16.6} & 396 ms & $\approx$3K more \\
    \bottomrule
  \end{tabular}
  \caption{Comparison with state-of-the-art activation functions using ILSVRC-2012 subset of the Imagenet~\cite{ILSVRC15}, and the standard branch of the Places2~\cite{zhou2016places}.}
  \vspace{-10pt}
  \label{table4}
\end{table}

\begin{figure*}[t]
\centering
    \begin{minipage}[c]{0.16\textwidth}
        \includegraphics[width=1\textwidth]{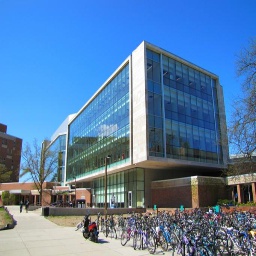}
    \end{minipage}
    \begin{minipage}[c]{0.16\textwidth}
        \includegraphics[width=1\textwidth]{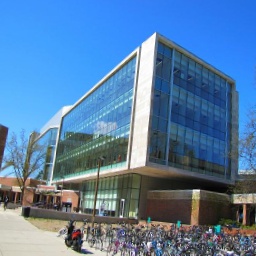}
    \end{minipage}
    \begin{minipage}[c]{0.16\textwidth}
        \includegraphics[width=1\textwidth]{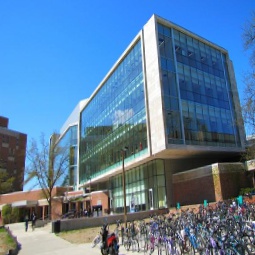}
    \end{minipage} 
    \begin{minipage}[c]{0.16\textwidth}
        \includegraphics[width=1\textwidth]{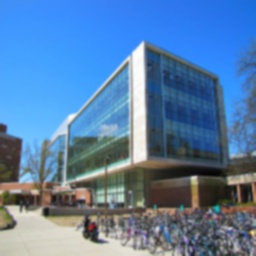}
    \end{minipage}
    \begin{minipage}[c]{0.16\textwidth}
        \includegraphics[width=1\textwidth]{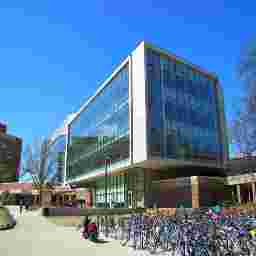}
    \end{minipage}
    \begin{minipage}[c]{0.16\textwidth}
        \includegraphics[width=1\textwidth]{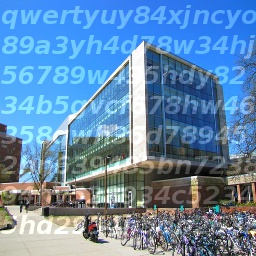}
    \end{minipage} 
    \\
    
    \begin{minipage}[c]{0.16\textwidth}
        \captionsetup[subfigure]{justification=centering}
        \centering
        \subcaption{Original image}
    \end{minipage}
    \begin{minipage}[c]{0.16\textwidth}
        \captionsetup[subfigure]{justification=centering}
        \centering
        \subcaption{Rotation}
    \end{minipage}
    \begin{minipage}[c]{0.16\textwidth}
        \captionsetup[subfigure]{justification=centering}
        \centering
        \subcaption{Perspective change}
    \end{minipage} 
    \begin{minipage}[c]{0.16\textwidth}
        \captionsetup[subfigure]{justification=centering}
        \centering
        \subcaption{Blur}
    \end{minipage}
    \begin{minipage}[c]{0.16\textwidth}
        \captionsetup[subfigure]{justification=centering}
        \centering
        \subcaption{Jpeg compression}
    \end{minipage}
    \begin{minipage}[c]{0.16\textwidth}
        \captionsetup[subfigure]{justification=centering}
        \centering
        \subcaption{In-painting}
    \end{minipage} 
    \vspace{-0.2cm}
    \\
    
    \begin{minipage}[c]{0.16\textwidth}
        \includegraphics[width=1\textwidth]{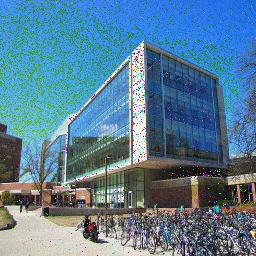}
    \end{minipage}
    \begin{minipage}[c]{0.16\textwidth}
        \includegraphics[width=1\textwidth]{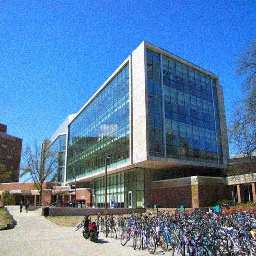}
    \end{minipage}
    \begin{minipage}[c]{0.16\textwidth}
        \includegraphics[width=1\textwidth]{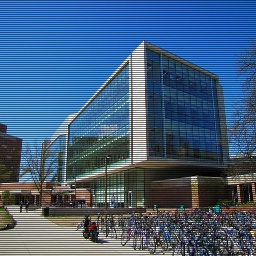}
    \end{minipage} 
    \begin{minipage}[c]{0.16\textwidth}
        \includegraphics[width=1\textwidth]{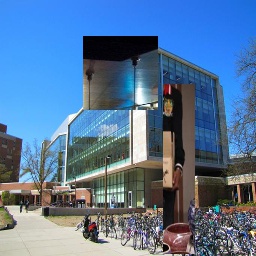}
    \end{minipage}
    \begin{minipage}[c]{0.16\textwidth}
        \includegraphics[width=1\textwidth]{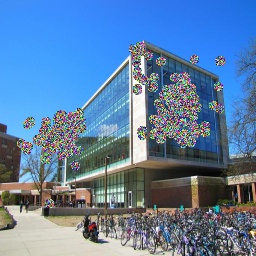}
    \end{minipage}
    \begin{minipage}[c]{0.16\textwidth}
        \includegraphics[width=1\textwidth]{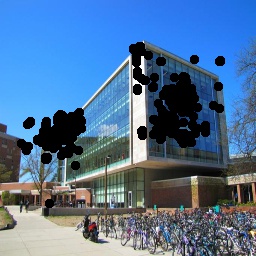}
    \end{minipage} 
    \\
    
    \begin{minipage}[c]{0.16\textwidth}
        \captionsetup[subfigure]{justification=centering}
        \centering
        \subcaption{Salt and pepper}
    \end{minipage}
    \begin{minipage}[c]{0.16\textwidth}
        \captionsetup[subfigure]{justification=centering}
        \centering
        \subcaption{Random noise}
    \end{minipage}
    \begin{minipage}[c]{0.16\textwidth}
        \captionsetup[subfigure]{justification=centering}
        \centering
        \subcaption{Structured noise}
    \end{minipage} 
    \begin{minipage}[c]{0.16\textwidth}
        \captionsetup[subfigure]{justification=centering}
        \centering
        \subcaption{Patch occlusion}
    \end{minipage}
    \begin{minipage}[c]{0.16\textwidth}
        \captionsetup[subfigure]{justification=centering}
        \centering
        \subcaption{Noise occlusion}
    \end{minipage}
    \begin{minipage}[c]{0.16\textwidth}
        \captionsetup[subfigure]{justification=centering}
        \centering
        \subcaption{Black occlusion}
    \end{minipage} 
    \vspace{-0.4cm}
\caption{Samples of deformed images obtained using different types of deformation.}
\vspace{-0.4cm}
\label{fig3}
\end{figure*}

\subsection{Experimental Analyses using Cifar, ILSVRC-12 and Places2 Datasets}
\label{subsec_cip}
In this section, we carry out a standard object classification task using benchmark datasets (CIFAR-10/100~\cite{krizhevsky2009learning}, ILSVRC-12~\cite{ILSVRC15}, Places2~\cite{zhou2016places}) to investigate the properties of our proposed method. For Cifar-10/100, we employ a VGG~\cite{simonyan2014very} style architecture by stacking many $3 \times 3$ convolution kernels (3 convolution layers in total). For ILSVRC-12 and Places2, we employ the same architecture with modification for multi-scale receptive fields, as used for our base model. We employ Batch Normalization (BN) together with ReLU activation functions after each convolution layer. Max-pooling operations endowed with $2 \times 2$ kernels are used for dimension reduction. For our proposed method, we replace all convolution layers to power convolutions with different configurations (we keep the Batch Normalization in CNNs if there was). We optimize these models by SGD with Nesterov’s accelerated gradient and momentum. Implementation details are provided in the supplemental material.

In the experiments employed using Cifar-10/100, we examine how the configuration of $\Lambda$ affects the performance of proposed power convolution method in CNNs. The average classification error and std. deviation (10 runs) are given in Table~\ref{table3}, boosted performances are observed for IN-ch. and OUT-ch. ($\Lambda=2$) configurations. However when the number of splits goes larger, the estimation of exponents becomes extremely difficult for OUT-ch. configurations, which results a degraded performance (Note that we could obtain the same results as base model (Base) by fixing exponents to be 1). We further compare the performance of the IN-ch. ($\Lambda=1, 2, 4$) using ResNet-36~\cite{he2015deep}, while performance boost is relatively trivial in Cifar-10, a gradually decreased error rate is observed in Cifar-100. We suspect that since in Cifar-100, number of samples for each class is 10 times fewer than that in Cifar-10, thus a dataset shift is more likely to happen in Cifar-100, which results a difference in performance.

In the experiments conducted using the ILSVRC-12 and Places2, we employ IN-ch. ($\Lambda=1, 2$) to examine the performance of our method. We also provide the computation cost (training time for each iteration) and complexity of networks (number of parameters). The results given in Table~\ref{table4} indicate the effectiveness of proposed method for large scale datasets, where the validation errors decrease by approximately 1\%--2\% using only about 0.04\% more parameters. Moreover, in order to demonstrate the potential of our proposed method, we further fine-tune a ResNet-152~\cite{he2015deep} model provided by the Places2, by replacing the convolution layers after each residual connection with IN-ch. ($\Lambda=1$) configuration. The fine-tuned model achieves a Top-1 and Top-5 classification accuracy of 56.1\% and 86.0\%, which improves the performance of the original model by 1.4\% and 1.0\%, respectively. To the best of our knowledge, by far this is the best performance obtained for a single model on the standard branch of Places2. 


\setlength{\tabcolsep}{4.8pt}
\begin{table*}[t]
  \centering
  \begin{tabular}{*{14}{c}}
    \toprule
    &
    \multirow{2}[0]{*}{\specialcell{\textbf{Models}}} & 
    \multirow{2}[0]{*}{\textbf{Rot.}} &
    \multirow{2}[0]{*}{\specialcell{\textbf{Perspe.}\\ \textbf{Change}}} & 
    \multirow{2}[0]{*}{\textbf{Blur}} &
    \multirow{2}[0]{*}{\specialcell{\textbf{Jpeg}\\ \textbf{Comp.}}} &
    \multirow{2}[0]{*}{\specialcell{\textbf{In-}\\ \textbf{Paint.}}} &    
    \multirow{2}[0]{*}{\specialcell{\textbf{Salt \&}\\ \textbf{Pepper}}} &
    \multirow{2}[0]{*}{\specialcell{\textbf{Rand.}\\ \textbf{Noise}}} &
    \multirow{2}[0]{*}{\specialcell{\textbf{Struc.}\\ \textbf{Noise}}} &
    \multirow{2}[0]{*}{\specialcell{\textbf{Patch}\\ \textbf{Occ.}}} & 
    \multirow{2}[0]{*}{\specialcell{\textbf{Tar. N.}\\ \textbf{Occ.}}} & 
    \multirow{2}[0]{*}{\specialcell{\textbf{Tar. B.}\\ \textbf{Occ.}}} &
    \multirow{2}[0]{*}{\textbf{Ave.}} \\
    \\
    \midrule
    \parbox[t]{2mm}{\multirow{3}{*}{\rotatebox[origin=c]{90}{\small{ILSVRC}}}}
    &
    ReLU & 
    9.2 & 
    13.0 & 
    17.4 & 
    18.5 & 
    55.3 & 
    46.6 & 
    17.1 & 
    39.5 & 
    15.3 & 
    14.0 & 
    17.5 & 
    24.0 \\
    &
    IN-ch. ($\Lambda=1$) & 
    8.4 & 
    12.3 & 
    19.3 & 
    18.7 & 
    54.2 & 
    51.1 & 
    16.6 & 
    36.7 & 
    14.5 & 
    13.7 & 
    \textbf{16.3} &
    23.8 \\
    &
    IN-ch.  ($\Lambda=2$) & 
    \textbf{7.4} &
    \textbf{10.8} &
    \textbf{17.0} &
    \textbf{16.5} &
    \textbf{51.9} &
    \textbf{40.9} &
    \textbf{13.6} &
    \textbf{19.7} &
    \textbf{14.0} &
    \textbf{13.6} &
    16.9 &
    \textbf{20.2} \\
    \midrule
    \parbox[t]{2mm}{\multirow{3}{*}{\rotatebox[origin=c]{90}{Places2}}}
    &
    ReLU & 
    16.8 & 
    13.6 & 
    31.8 & 
    14.6 & 
    48.5 & 
    53.8 & 
    23.2 & 
    56.7 & 
    9.7 & 
    12.2 & 
    14.2 & 
    26.8 \\
    &
    IN-ch. ($\Lambda=1$) & 
    16.2 & 
    12.7 & 
    27.4 & 
    15.2 & 
    46.1 & 
    51.1 & 
    21.1 & 
    55.1 & 
    9.6 & 
    11.1 & 
    13.8 & 
    25.4 \\
    &
    IN-ch. ($\Lambda=2$) & 
    \textbf{14.5} &
    \textbf{11.1} &
    \textbf{25.1} &
    \textbf{12.4} &
    \textbf{44.5} &
    \textbf{45.6} &
    \textbf{17.3} &
    \textbf{52.4} &
    \textbf{8.9} &
    \textbf{10.5} &
    \textbf{11.7} &
    \textbf{23.1} \\
    \bottomrule
  \end{tabular}
  \caption{Classification error (Top-1 error(\%)) obtained using deformed images. Results given in top 3 and bottom 3 rows show the results obtained using ILSVRC-12 and Places2 datasets respectively.}
  \label{table5}
  \vspace{-0.45cm}
\end{table*}

\subsection{Analysis of Robustness of Learned Deep Features to Image Deformation}
\label{subsect_robust}
In this section, we examine the robustness of models obtained in the previous section using images deformed by various deformation models. Here, by the term ``deformation'', we do not only refer to arbitrary geometric transformations applied to images, such as rotation, scaling, translation, and other affine transforms~\cite{ImageProcessing}, but also consider several types of deformations that are widely observed in practice. We categorize the latter deformations into statistical noises, signal processing loss, and occlusions. Samples of deformed images obtained using different deformation methods are given in Figure~\ref{fig3}. More detailed algorithms used for generating the deformed images are provided in the supplemental material.

\noindent \textbf{Geometric Transformations:} Geometric transformations are the deformations most considered for CNNs. Data augmentation approaches such as scaling and translation are commonly carried out in various state-of-the-art models~\cite{goodfellow13, krizhevsky2012imagenet, szegedy2015going}. We also employ the same augmentation methods for training models in the previous section, hence we mainly consider rotation and perspective transformation in this section. Concretely, for rotation, we rotate images with a small angle up to $\pm 15^{\circ}$ while leaving the scale unchanged. For perspective transformation~\cite{Carroll:2010:IWA:1778765.1778864}, we arbitrarily choose four pixels within the image as the base pixels of new perspective. Then, we apply the corresponding affine transform with respect to the base pixels to generate deformed images.

\noindent \textbf{Signal Processing Loss:} We consider this type of deformation as information loss occurred during processing of 2D images. We choose two specific cases for generating deformed images, i) the Gaussian blur and ii) the Jpeg compression~\cite{wallace1992jpeg}. Applying a Gaussian blur on an image attenuates the image's high-frequency components, hence the information in the corresponding frequency is lost. To generate the blurred images, we convolve the image with a 2D Gaussian blur kernel of size $5\times5$. The Jpeg compression is a popularly used image compression method that offers a selectable trade-off between storage size and image quality. Encoding steps of Jpeg compression such as down-sampling and quantization will result in certain loss of information. Especially when a large compression ratio is employed, severe high frequency loss can be observed. 

\noindent \textbf{Statistical Noise:} In this work, we focus on three different types of noise with particular statistical properties, that can be widely observed in digital images; i) random noise (non-Gaussian), ii) salt and pepper noise (impulse valued noise)~\cite{boyat2015review}, and iii) structured noise. Random noise is characterized by intensity and color fluctuations above and below actual image intensity. Salt and pepper noise randomly drops original values of some pixels according to a distribution, instead of corrupting the whole image. We sample the dropped values from a Gaussian distribution with a large standard deviation and corrupt up to 20\% of total pixels to make the noise more aggressive. Structured noise is introduced by interference of electronic devices with high spatial dependency~\cite{boyat2015review} in practice. Here we employ a simple structure that corrupting every second row of an image matrix. 

\noindent \textbf{Occlusions:} We also deform images using occlusion methods, which may cause essential disruption for scene recognition using CNNs. We consider three different types of artificial occlusions; i) in-painting~\cite{Bertalmio:2000:II:344779.344972}, ii) attention targeted occlusion~\cite{sun2015design} and iii) semantic patch occlusion. For in-painting, we employ randomly generated strings with different transparency. Attention targeted occlusion is designed to obliterate the information \textit{important} for recognition of a target class. Concretely, a saliency map that records pixel-wise classification scores is calculated using gradient methods. Then, we occlude some clusters of pixels with noise or black masks within the areas that contribute most to the final classification score. For semantic patch occlusion, we cover parts of the target images with patches randomly cropped from other images. The size of the patches is set to be larger than 1/64 of original image size to for including semantic information.

The robustness is evaluated by Top-1 and Top-5 classification error for deformed images. In order to obtain comparable results, during generation of test datasets, we \textit{only} apply those images that are correctly classified (Top-1 guess) by \textit{all} the tested models. In total, approximately 25,000 and 18,000 images obtained from validation sets of ILSVRC-12 and Places2 are used for data generation, respectively. Detailed results obtained for different models and different deformations are given in Table~\ref{table5}. The results show that the model equipped with the proposed methods outperform the base model on both datasets in general. IN-ch. ($\Lambda=2$) provides 3.8\% and 3.7\% (15.8\% and 13.8\% relatively) performance boost in average. Especially, for deformation patterns such in-painting, random noise and structured noise, the error rates are decreased by 4\%--20\%. On the other hand, average boosts of IN-ch. ($\Lambda=1$) model are 0.2\% and 1.4\%. Interestingly, for blur and Salt \& Pepper noise, the method fails in the classification task of single object while obtaining better robustness using scene images, which are relatively complicated with respect to the amount of contents in images. Similarly, the boosts in the targeted occlusions on ILSVRC-12 are not obvious for proposed methods as well, while decent boosts have been achieved on Places2. We suspect that lack of information on spatial distribution is likely to result in the degraded performance for the classification decision of single object. Meanwhile, scenes usually contain more information on spatial distributions for objects and patterns that can be better captured by proposed methods, thus a consistent decrement of error is found.

\setlength{\tabcolsep}{4.2pt}
\begin{table}[t]
  \centering
  \begin{tabular}{cccccc}
    \toprule
    \textbf{Models} & \textbf{Ori.} & \textbf{Comp.} & \textbf{Blur} & \textbf{S.\&P.} & \textbf{Stru.} \\
    \midrule
    ZF~\cite{zeiler2014visualizing, ren2015faster} & \textbf{58.7} & 54.9 & 43.7 & 45.1 & 14.5 \\ 
    ZF-IN-ch. & \textbf{58.7} & 55.8 & 44.8 & 47.2 & 17.3 \\ 
    \midrule
    \midrule
    \textbf{Models} & \textbf{In-P.} & \textbf{Tar.B.} & \textbf{Tar.N.} & \textbf{Mix.L.} & \textbf{Mix.H.} \\
    \midrule
    ZF & 43.5 & 42.7 & 42.8 & 50.0 & 14.3 \\ 
    ZF-IN-ch. & 45.3 & 43.8 & 44.6 & 52.2 & 17.6 \\ 
    \bottomrule
  \end{tabular}
  \caption{Detection performance (\% mAP) for Pascal Voc 2007 validation set with different patterns of deformation.}
  \vspace{-20pt}
  \label{table6}
\end{table}

\begin{figure*}[t]
\centering
    \begin{minipage}[c]{0.233\textwidth}
        \includegraphics[width=1\textwidth]{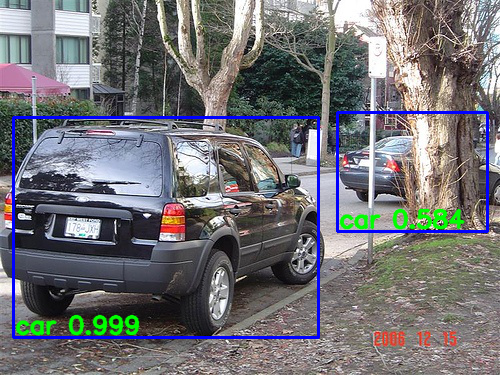}
    \end{minipage}
    \begin{minipage}[c]{0.233\textwidth}
        \includegraphics[width=1\textwidth]{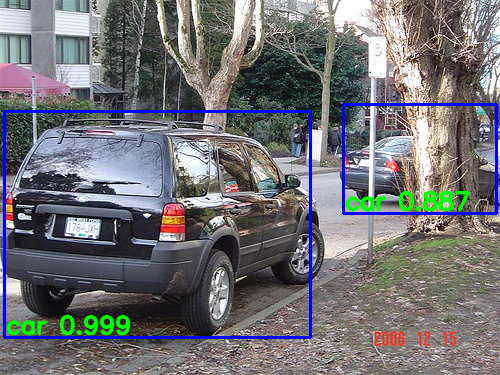}
    \end{minipage}
    \begin{minipage}[c]{0.233\textwidth}
        \includegraphics[width=1\textwidth]{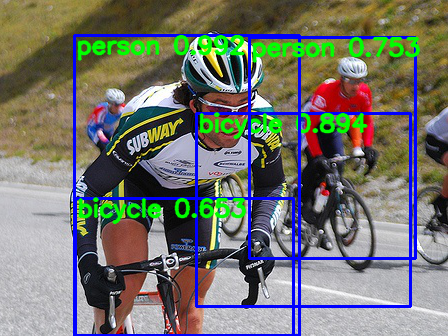}
    \end{minipage} 
    \begin{minipage}[c]{0.233\textwidth}
        \includegraphics[width=1\textwidth]{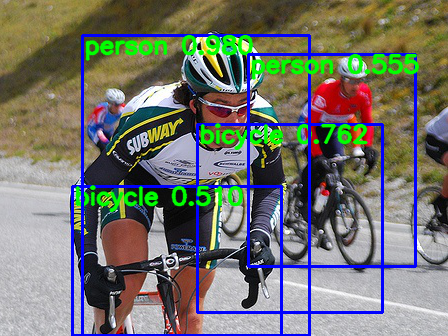}
    \end{minipage} 
    \\

\centering
    \begin{minipage}[c]{0.233\textwidth}
        \includegraphics[width=1\textwidth]{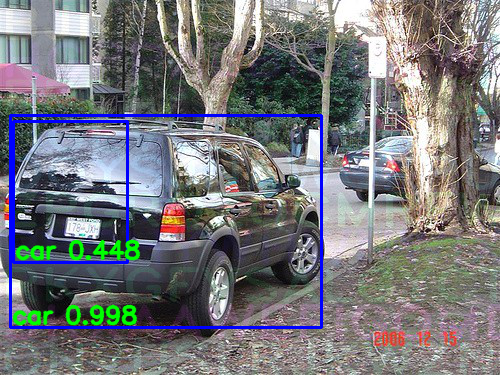}
    \end{minipage}
    \begin{minipage}[c]{0.233\textwidth}
        \includegraphics[width=1\textwidth]{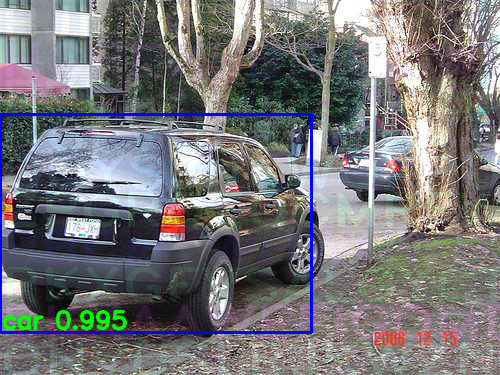}
    \end{minipage}
    \begin{minipage}[c]{0.233\textwidth}
        \includegraphics[width=1\textwidth]{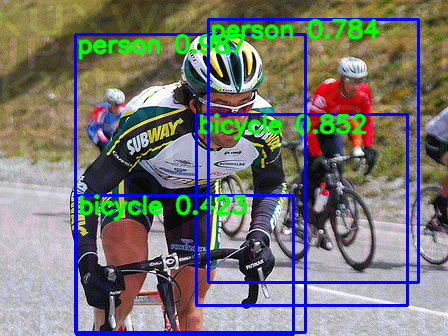}
    \end{minipage} 
    \begin{minipage}[c]{0.233\textwidth}
        \includegraphics[width=1\textwidth]{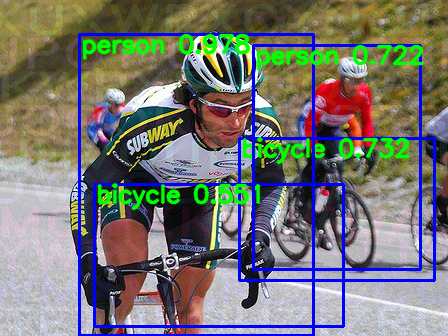}
    \end{minipage} 
    \\

\centering
    \begin{minipage}[c]{0.233\textwidth}
        \includegraphics[width=1\textwidth]{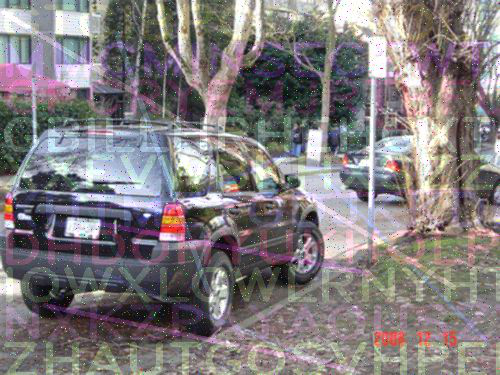}
    \end{minipage}
    \begin{minipage}[c]{0.233\textwidth}
        \includegraphics[width=1\textwidth]{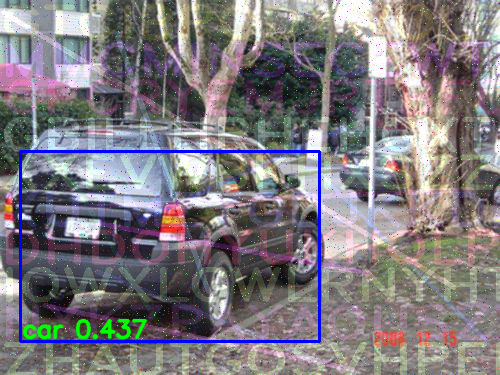}
    \end{minipage}
    \begin{minipage}[c]{0.233\textwidth}
        \includegraphics[width=1\textwidth]{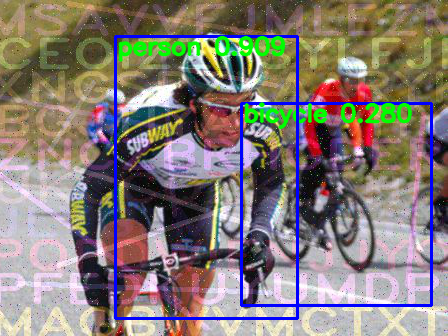}
    \end{minipage} 
    \begin{minipage}[c]{0.233\textwidth}
        \includegraphics[width=1\textwidth]{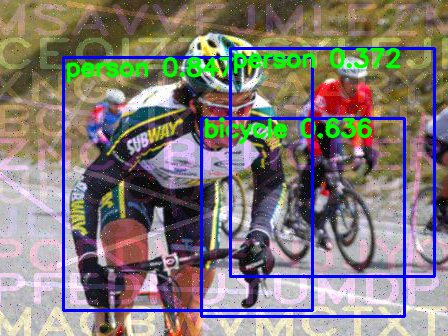}
    \end{minipage} 
    \\

\centering
    \begin{minipage}[c]{0.233\textwidth}
        \captionsetup[subfigure]{justification=centering}
        \centering
        \subcaption*{\textbf{ZF}}
    \end{minipage}
    \begin{minipage}[c]{0.233\textwidth}
        \captionsetup[subfigure]{justification=centering}
        \centering
        \subcaption*{\textbf{ZF-IN-ch. ($\Lambda=1$)}}
    \end{minipage}
    \begin{minipage}[c]{0.233\textwidth}
        \captionsetup[subfigure]{justification=centering}
        \centering
        \subcaption*{\textbf{ZF}}
    \end{minipage}
    \begin{minipage}[c]{0.233\textwidth}
        \captionsetup[subfigure]{justification=centering}
        \centering
        \subcaption*{\textbf{ZF-IN-ch. ($\Lambda=1$)}}
    \end{minipage}
    \vspace{-0.4cm}
    \\
\caption{Examples of objects detection results obtained using deformed images. Text given in green color indicates the class of objects with confidence. Rows from top to bottom: Original images, images with light deformation (In-Painting, Gaussian Noise), images with heavy deformation (In-Painting, Gaussian Noise, Salt and Pepper Noise, Blurring, JPEG Compression). Left and right images are selected from \text{training} and validation set, respectively. }
\label{fig4}
\vspace{-0.25cm}
\end{figure*}

\subsection{Analyses of Detection Results using the Pascal Voc 2007 Dataset}
\label{detection}
In this section, we examine the performance of our proposed method for object detection task using the Pascal Voc 2007 dataset. We employ Faster-RCNN~\cite{ren2015faster} as the detection algorithm, and Zeiler and Fergus (ZF) model~\cite{zeiler2014visualizing} as the baseline CNN model. Then, we append our proposed IN-ch. ($\Lambda=1$) to the last three convolution layers of ZF model (ZF-IN-ch. ($\Lambda=1$)), and evaluate the change in performance. In order to ensure the fairness for evaluation of detection, we implement both ZF and ZF-IN-ch. ($\Lambda=1$) models for training using the ILSVRC-2012 with random initialization. Then, we manually select a snapshot of the ZF-IN-ch. ($\Lambda=1$) model which provides classification accuracy (Top-1/5 58.6\%/81.7\%) that is same to that of the fully trained ZF model. Then, we train both models using the Pascal Voc 2007 training dataset, and we manually select a snapshot of ZF-IN-ch. ($\Lambda=1$) model which provides the detection performance (58.7\% mAP) that is same to that of the fully trained ZF model. We employ the deformation patterns given in Section~\ref{subsect_robust}, and two datasets (Mix.Light and Mix.Heavy) that employ mixed patterns of deformation (see Figure~\ref{fig4}). The results given in Table~\ref{table6} show that, although both models have the same detection performance in the original validation set, the model equipped with IN-ch. ($\Lambda=1$) power convolution gains approximately 1\%--3\% mAP under different deformations. Figure~\ref{fig4} provides some examples of detection results.



\section{Conclusions and Discussions}
In this work, we aim to enhance the robustness of CNNs to image deformation for popular object recognition and detection problems. We consider this challenge as a dataset shift problem, where the higher moment statistics of feature distributions shift due to deformation. In order to attenuate this effect, we employ power operations with learnable parameters as non-linearity, and combine them into convolution operation in CNNs. We give insights into the efficiency of our proposed method in dealing with dataset shift problem, compared to other different types of non-linearity. The experimental results obtained using benchmark datasets indicate a substantial boost of robustness of feature representations to various types of deformations. We believe that this approach can be beneficial for training of CNNs in various computer vision tasks, where deformations may impair the performance, such as object identification and detection, image retrieval and restoration.

\vspace{0.5cm}
{\small
\bibliographystyle{ieee}
\bibliography{iccv_2017}
}

\end{document}